\documentclass[final,3p,times,twocolumn]{elsarticle}

\usepackage{amssymb}
\usepackage{amsmath}
\usepackage{amsthm}
\usepackage{acro}
\usepackage{graphicx}
\usepackage{mathtools}
\usepackage{tikz}
\usetikzlibrary{chains,fit,shapes.geometric, arrows.meta}
\usepackage{multirow}
\usepackage{algorithm2e}
\usepackage{booktabs}
\usepackage{smartdiagram}
\usepackage{hyperref}
\usepackage{pgfplots}
\usepackage{svg}
\usepackage{makecell}

\theoremstyle{remark}

\tikzset{%
  every neuron/.style={
    circle,
    draw,
    minimum size=1cm
  },
  neuron missing/.style={
    draw=none, 
    scale=4,
    text height=0.333cm,
    execute at begin node=\color{black}$\vdots$
  },
}

\usepackage{amssymb}
\usepackage{svg}

\journal{Journal of Web Semantics}
\acsetup{make-links = true, first-style = short-long, list/display=all, list/template=toc, list/sort = false , list/locale/display = true, locale/format=\emph}

\DeclareAcronym{ai}{
  short = Artificial Intelligence ,
  long = {AI}
}
\DeclareAcronym{ann}{
  short = Artificial Neural Network ,
  long = {ANN}
}
\DeclareAcronym{c}{
  short = Clustering ,
  long = {C}
}
\DeclareAcronym{cai}{
  short = Comprehensible Artificial Intelligence ,
  long = {CAI}
}
\DeclareAcronym{gnn}{
  short = Graph Neural Network ,
  long = {GNN}
}
\DeclareAcronym{gc}{
  short = Graph Clustering ,
  long = {GC}
}
\DeclareAcronym{iml}{
  short = Interpretable Machine Learning ,
  long = {IML}
}
\DeclareAcronym{iid}{
  short = Independent and Identically Distributed ,
  long = {IID}
}
\DeclareAcronym{kg}{
  short = Knowledge Graph ,
  long = {KG}
}
\DeclareAcronym{kaw}{
  short = Knowledge Aware ,
  long = {KAW}
}
\DeclareAcronym{kaq}{
  short = Knowledge Acquisition ,
  long = {KAQ}
}
\DeclareAcronym{kge}{
  short = Knowledge Graph Embedding ,
  long = {KGE}
}

\DeclareAcronym{ilp}{
  short = Inductive Logic Programming ,
  long = {ILP}
}
\DeclareAcronym{lrp}{
  short = Layer-wise Relevance Propagation ,
  long = {LRP}
}
\DeclareAcronym{ml}{
  short = Machine Learning ,
  long = {ML}
}
\DeclareAcronym{nlp}{
  short = Natural Language Processing ,
  long = {NLP}
}
\DeclareAcronym{pgm}{
  short = Probabilistic Graphical Model ,
  long = {PGM}
}
\DeclareAcronym{rnn}{
  short = Recursive Neural Network,
  long = {RNN}
}
\DeclareAcronym{rbl}{
  short = Rule-Based Learning,
  long = {RBL}
}
\DeclareAcronym{rl}{
  short = Reinforcement Learning ,
  long = {RL}
}
\DeclareAcronym{xai}{
  short = Explainable Artificial Intelligence ,
  long = {XAI}
}
\DeclareAcronym{fm}{
  short = Factorization Machine ,
  long = {FM}
}

\DeclareAcronym{lp}{
  short = Link Prediction ,
  long = {LP}
}
\DeclareAcronym{nc}{
  short = Node Classification ,
  long = {NC}
}
\DeclareAcronym{owl}{
  short = Web Ontology Language ,
  long = {OWL}
}
\DeclareAcronym{r}{
  short = Recommendation ,
  long = {R}
}
\DeclareAcronym{pf}{
  short = Pathfinding ,
  long = {PF}
}
\DeclareAcronym{tl}{
  short = Translational Learning,
  long = {TL}
}
\begin{document}
\onecolumn
\twocolumn
\begin{frontmatter}



\title{Comprehensible Artificial Intelligence on Knowledge Graphs: A Survey.}
%


\author[inst1,inst2]{Simon Schramm\fnref{label1}} 
\author[inst1]{Christoph Wehner\fnref{label1}}
\fntext[label1]{The authors contributed equally.}
\author[inst1]{Ute Schmid}

\affiliation[inst1]{organization={Cognitive Systems Group, University of Bamberg},
            addressline={An der Weberei 5}, 
            city={Bamberg},
            postcode={96049}, 
            state={Bavaria},
            country={Germany}}
\affiliation[inst2]{organization={BMW Group},
            addressline={Petuelring 130}, 
            city={Munich},
            postcode={80809}, 
            state={Bavaria},
            country={Germany}}

\begin{abstract}
Artificial Intelligence applications gradually move outside the safe walls of research labs and invade our daily lives. This is also true for Machine Learning methods on Knowledge Graphs, which has led to a steady increase in their application since the beginning of the $21st$ century. 
However, in many applications, users require an explanation of the AI's decision. This led to increased demand for Comprehensible Artificial Intelligence. Knowledge Graphs epitomize fertile soil for Comprehensible Artificial Intelligence, due to their ability to display connected data, i.e. knowledge, in a human- as well as machine-readable way. 
This survey gives a short history of Comprehensible Artificial Intelligence on Knowledge Graphs. Furthermore, we contribute by arguing that the concept of Explainable Artificial Intelligence is overloaded and overlaps with Interpretable Machine Learning. By introducing the parent concept Comprehensible Artificial Intelligence, we provide a clear-cut distinction of both concepts while accounting for their similarities. Thus, we provide in this survey a case for Comprehensible Artificial Intelligence on Knowledge Graphs consisting of Interpretable Machine Learning and Explainable Artificial Intelligence on Knowledge Graphs. This leads to the introduction of a novel taxonomy for Comprehensible Artificial Intelligence on Knowledge Graphs. In addition, a comprehensive overview of the research in this research field is presented and put into the context of the taxonomy.
Finally, research gaps in the field are identified for future research. 

\end{abstract}





\end{frontmatter}




\section{Overview}\label{sec:ov}
Artificial Intelligence (AI) applications gradually move outside the safe walls of research labs and invade our daily lives. Many real-world AI applications, e.g., in medicine \citep{Bruckert2020}, industrial product development \citep{Schramm2022, Wehner2023}, and autonomous driving \citep{Wehner2022}, are safety-critical. Thus, AI algorithms used in those fields are regulated \citep{eu2021}. The regulations aim to give the user the ability to comprehend the AIs decision \citep{eu2021}. However, most state-of-the-art AI methods, like deep neural networks, are too complex and complicated to comprehend for a user, so-called black box models. Leading to increased demand for Comprehensible Artificial Intelligence (CAI) research in the last years \citep{Schwalbe2021, Futia2020}. 

CAI is a set of methods that enable stakeholders to understand and retrace the output of AI models \citep{Rudin2019, Schmid2021a, Futia2020}. CAI has two major sides. One side reconstructs the decision-making process of a black box model with the help of a human-understandable mapping from the model's input to its output. This side is called Explainable Artificial Intelligence (XAI) \citep{Rudin2019}. 
The other side of CAI aims to create AI models with, by default, human-understandable decision-making processes, so-called white boxes. This side is called Interpretable Machine Learning (IML) \citep{Rudin2019}. Section \ref{sub:meth} provides an in-depth discussion of CAI. 

\begin{figure}[!ht]
    \centering
    \includegraphics[width=.45\textwidth]{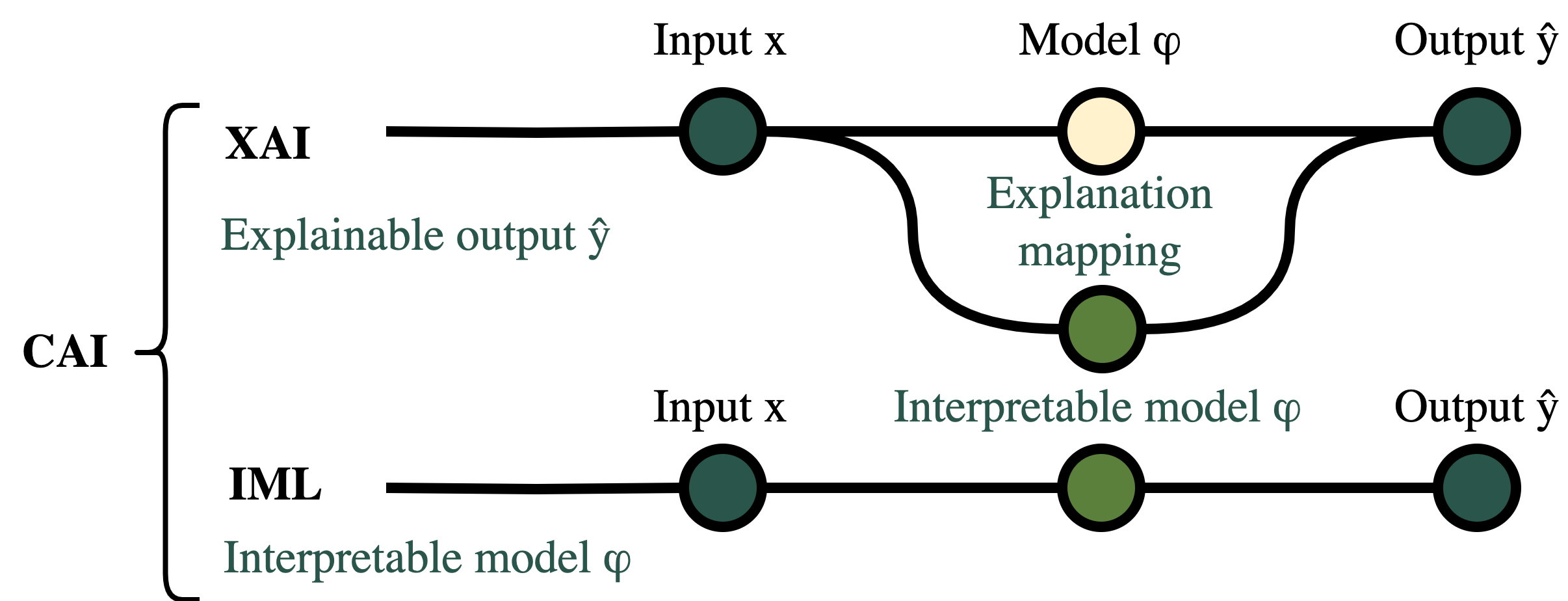}
    \caption{Generic CAI framework.}\label{fig:cai}
\end{figure}

\paragraph{On the necessity of Comprehensible Artificial Intelligence on Knowledge Graphs}
Regularly, AI models with human-understandable decision-making processes are regarded as \a. However, the concept XAI is overloaded. For example, take \emph{TEM} \citep{Wang2018} and \emph{RelEx} \citep{Zhang2021}. Both methods talk about explainability. While both satisfy similar high-level needs, e.g., justifying the model behavior, identifying risks, debugging the model, and uncovering unknown rules, they are fundamentally different. \\
One method (\emph{TEM}) holds a complete Machine Learning (ML) model, with inherent interpretability by human-understandable inner mechanics. \\
The other method (\emph{RelEx}) is a post-hoc explainability method applied to a black box model to discover which part of the input contributed to the output. \\
We aim to account for their differences by taxonomizing such methods in IML and XAI. And at the same time, we acknowledge their similarities in taxonomizing IML and XAI as the two components of CAI (cf. Figure \ref{fig:cai}). 

Knowledge Graphs (KG) are human- and machine-readable representations of semantically linked data, referred to as knowledge over a particular domain \citep{Hogan_2022}. Semantically linked data is predestined to create AI models with human-understandable decision-making processes \citep{Gaur2021, Futia2020}, which shall be the notion of CAI in this survey. This survey aims to give an overview of CAI methods on KGs.  

We start with an introduction to the history of CAI on KGs. Thereby, we define the term CAI and put the notions IML and XAI into context. Furthermore, we taxonomize state-of-the-art approaches to CAI on KGs by representation, task, foundation, and comprehensibility. Finally, opportunities for future research of CAI on KGs are laid out. The following section provides a brief history of CAI on KGs, followed by the proceeding and sources of the survey, the scope of the survey, and the related work.

\subsection{A brief history of Comprehensible Artificial Intelligence on Knowledge Graphs}\label{sub_hist}
The first KGs were created as semantic networks for a university project from 1972 to 1980 \citep{Schneider1973CourseMA}. Between 1985 and 2007 the idea of a KG was applied to different fields, such as human language in \emph{Wordnet} \citep{Miller1995}. Algorithms mainly focused on utilizing the symbolic semantics of the KG to reason and learn new rules over the domain of a KG \citep{Muggleton1991, perez2006}. These algorithms, like \emph{FOIL} \citep{Quinlan1990}, were fully comprehensible. The success of KGs rise abruptly and rapidly when \emph{DBpedia} \citep{Auer2007} and \emph{Freebase} \citep{Bollacker2008} were created as general-purpose KGs, which led to the \emph{Google} KG in 2012 \citep{singhal_2012}. Since then, a plethora of private companies and academic institutions utilize KGs for a variety of applications \citep{Hogan_2022}. 

Researchers introduced a wide range of ML methods operating on KGs, fueled by the ever-growing interest in KGs. This gave rise in the last decade to incomprehensible methods, like translational models \citep{Bordes2013} and Graph Neural Networks (GNN) \citep{scarselliGraphNeuralNetwork2009}. However, the incomprehensibility of those algorithms prohibits their applications in high-stake applications. Thus, recently CAI \citep{Futia2020} KGs has gained significant attention, which is also the focus of this survey. 
 
 
\subsection{Proceeding and sources}\label{sub:proc}

Following the systematic literature search process of \citet{websterAnalyzingPrepareFuture2002} and \citet{vombrockeReconstructingGiantImportance2009}, we identified 120 researchers from the community publishing papers in the domain of CAI and KGs.

\begin{figure*}[!ht]
    \centering
    \includegraphics[scale=0.4]{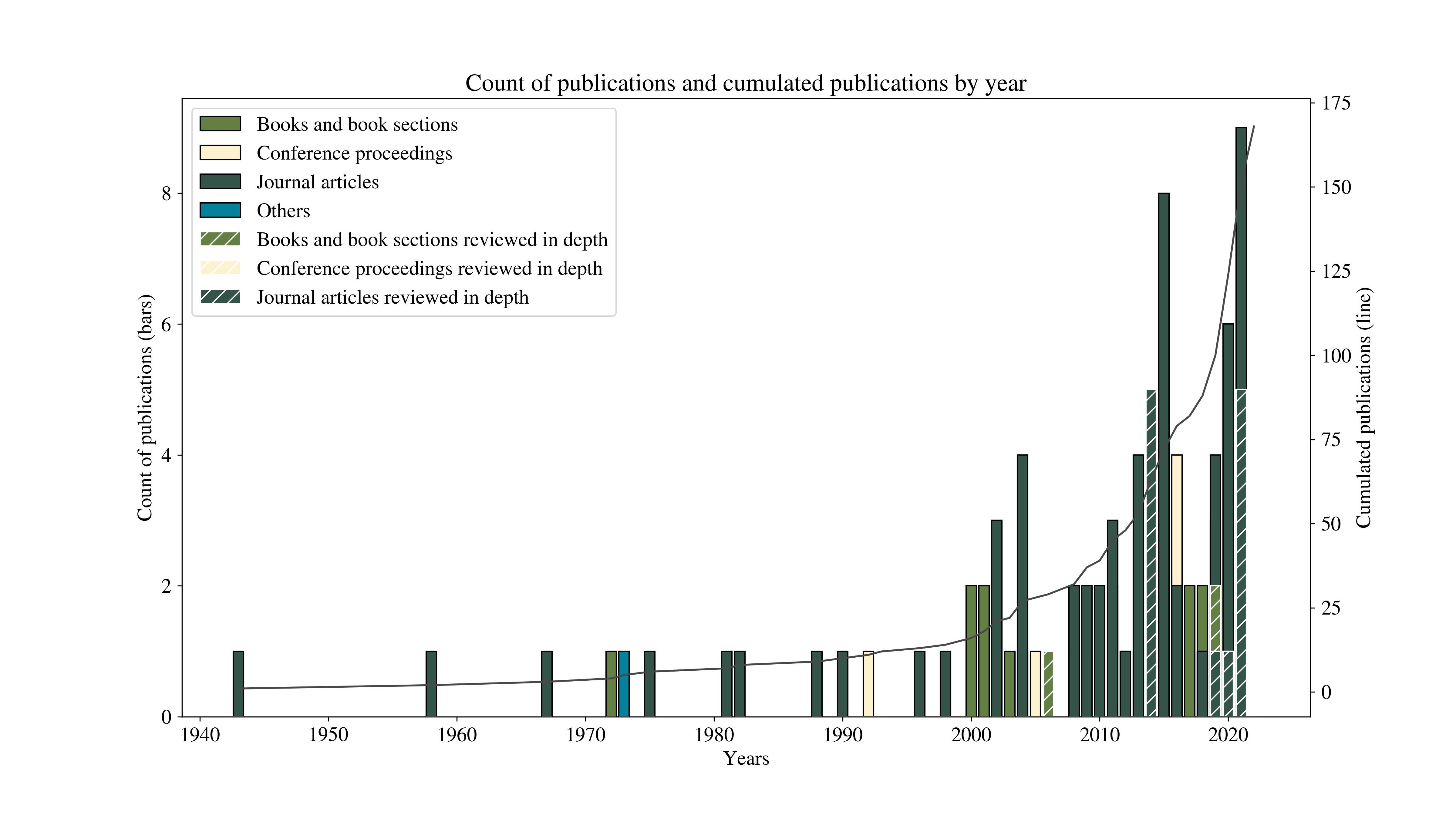}
    \caption{Reviewed publications by year.}\label{fig:pub_by_year}
\end{figure*}

We contacted our fellow researchers posing the following questions: \\
(1) Which papers do you think are crucial to cite in such a survey? \\
(2) Which major outlets (conferences/journals) have you seen published literature on Comprehensible Artificial Intelligence on Knowledge Graphs? \\
(3) Are there topics you would like to see discussed in such a survey? \\
(4) Where do you see future research directions for Comprehensible Artificial Intelligence on Knowledge Graphs? \\
The answers provided to these questions can be found in \ref{sec:appendix_a}. As a result of this questionnaire, we determined the search string \emph{"Knowledge Graph*" \textbf{AND} "Machine Learning*" \textbf{OR} "Artificial Intelligence*"} and \emph{"Knowledge Graph*" \textbf{AND} "Interpretable*" \textbf{OR} "Explainable*" \textbf{OR} "Comprehensible*"}, which we searched for in titles, abstracts, and keywords. Figure~\ref{fig:pub_by_year} displays the resulting 163 publications. Figure~\ref{fig:pub_by_year} reveals, that especially journal articles accumulate in the very recent past.


We utilized the \emph{Scimago Institutions Rankings}\footnote{\emph{Scimago Institutions Rankings} is a ranking of scientific journals and conferences, based on a composite indicator that represents research performance, innovation outputs and societal impact measured by the web visibility of the respective journal or conference \citep{scimagoresearchgroupOECDSCImagoResearch2014, bornmannWhatProportionExcellent2014, bornmannNewExcellenceIndicator2012}.}, to assure a high quality in selecting publications.
We focused on journals and conference proceedings above rank 150. However, some articles we cite are not published yet and are thus only available as preprint. Table~\ref{tab:pub_freq} presents the most frequent journals, conferences or publishers of the literature list of this survey. 
Eventually, we chose 55
publications from a total of 163 titles obtained by our structured search, based on an assessment of the publications' abstracts. In the following sections, the 55 publications chosen for close assessment will be described, and taxonmized according to Table~\ref{tab:scop}. Later, a combination of the taxonomy provided by this survey with the discrimination of IML and XAI on KGs will be provided in Figure~\ref{fig:heatmpa}.

\begin{table}[!ht]
        \centering
        \footnotesize
        \begin{tabular}{lrrr}
        \toprule
        Type                                                    &  Book             &  Conference   &  Journal       \\
                                                                &  (section)        &  proceeding   &  article       \\
        Publisher                                               &                   &               &                \\
        \midrule
        Springer                                                &2                  &8              &2                \\
        IEEE&0                  &0              &1                 \\
        Elsevier                                                &0                  &0              &2               \\
         ACM               &0                  &13             &0                \\
        Others                                                  &0                  &10              &17                \\
        \bottomrule
        \end{tabular}
        \normalsize
        \caption{Most frequent publishers by publication type.}\label{tab:pub_freq}
\end{table}

The following Section~\ref{sub:scop} will outline the scope and the taxonomy of this survey. 
\subsection{Scope of the survey}\label{sub:scop}


Based on the typology of literature reviews developed by \citet{pareSynthesizingInformationSystems2015}, the survey at hand aims at a narrative method to describe the state of the art of CAI on KGs. The motivation for this scope arises from the surging interest in KGs as input for ML methods and the mounting importance of comprehensibility in AI. Thereby, the survey follows the notion of CAI as a conglomerate of methods that reconstruct the decision-making process of a black box AI model with the help of a human-understandable mapping from the model's input to its output \textbf{and} methods that make the decision-making process of an AI model human-understandable \cite{molnarInterpretableMachineLearning2019, roscherExplainableMachineLearning2020, murdochDefinitionsMethodsApplications2019, liptonMythosModelInterpretability2018}.
The concept CAI is necessary, as there has yet to be a clear-cut distinction between IML and XAI in past publications.
Methods for human-understandable AI are generally referred to as XAI by previous publications. However, this leaves ML methods with inherent interpretability and post-hoc explanation methods in the same class. 
This survey aims to clarify the distinction between the concepts Interpretable Machine Learning and Explainable Artificial Intelligence, as well as to define the unifying term CAI. Since this scope entails both systems that learn explanations and systems that do not involve a learning process, it makes sense to refer to comprehensibility in the context AI, rather than ML. Hence, in Section~\ref{sub:comp}, we will come to such a formal distinction, while summarizing both concepts by the notion of CAI. 

\begin{figure}[!ht]
    \centering
    \includegraphics[scale=0.35]{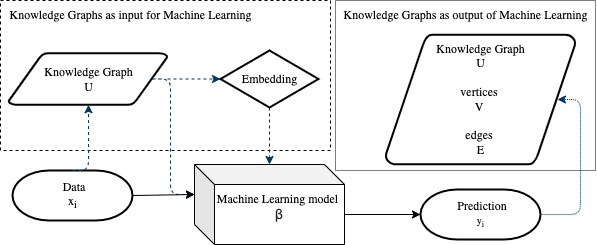}
    \caption{Simplified and generic KG-ML pipeline with KGs as input or output.}\label{fig:scope}
\end{figure}

In the context of CAI, a KG $U$ can be used as input for ML, whereby data $x_i$ forms the KG $U$ that can be embedded for being used in a ML model $\beta$. Other than that, KGs can also be defined as target output representations of an algorithm.\footnote{Recent progress in computer vision serves as an vivid example, elaborating on techniques that correctly infer relations between detected objects, such as, \emph{window} \emph{on} \emph{building}, or \emph{building} \emph{near} \emph{person} \citep{xuReasoningRCNNUnifyingAdaptive2019}.} Figure~\ref{fig:scope} visualizes this simplified, generic KG-ML pipeline. 



KGs as target output representations of a model $\beta$ cannot render the model interpretable, nor can they epitomize an output that refers to the input data $x_i$ in order to generate explainability. Henceforth, the focus of this survey is CAI on KGs as an input for ML. 



In accordance with \citet{websterAnalyzingPrepareFuture2002}, 
we developed a taxonomy for the publications reviewed, based on the kind of representation, the setting of the KG, the kind of optimization, the task of the AI model and the type of comprehensibility that is underlying. The taxonomy is shown in Table \ref{tab:scop}. A profound description of the scopes of our taxonomy will be provided in Section~\ref{sub:meth}.



\begin{table*}[!ht]
\centering
\footnotesize
    \begin{tabular}{l|p{0.7\linewidth}}
    \toprule
    \textbf{Scope}                  & \textbf{Characterization} \\
    \toprule
    Representation      & Symbolic (SB), Sub-symbolic (SSB), Neuro-symbolic (NSB). \\ 
    \midrule
    Foundation        & Factorization Machines (FM), Translational Learning                          (TL), Rule-based Learning (RBL),  \\
                        & Neural Networks (ANN), Reinforcement Learning (RL), Others (O).  \\
    \midrule
    Task                & Link Prediction (LP), Node Clustering (NC), \\ 
                        & Graph Clustering (GC), Clustering (C), Recommendation (R).  \\
    \midrule                   
    Comprehensibility   & Interpretable Machine Learning (IML), \\
                        & Explainable Artificial Intelligence (XAI).\\
    \bottomrule
    \end{tabular}
    \normalsize
\caption{Taxonomy of the survey.}
\label{tab:scop}
\end{table*}

\subsection{Related work} \label{sub:related_work}
Several surveys exist related to CAI and KGs, although they usually refer to XAI instead of CAI. \\
For example, \citet{Tiddi2021} elaborate on CAI on KGs. They follow a broader understanding of CAI. \citet{Tiddi2021} consider methods as CAI that rationalize the decision of an AI model through external reasoning, independent of the AI model's decision-making process. In contrast, the survey at hand understands CAI solely as a method of transparent decision-making AI models. \\
\citet{Bianchi2020} apply a similar understanding of CAI to the survey at hand. However, \citet{Bianchi2020}`s survey focuses on a general overview of AI with KGs as an input, including comprehensible methods. In contrast, the survey at hand fully commits to CAI on KGs. \\
In addition, \citet{Lecue2020} elaborate on CAI on KGs. They structure their paper by the AAAI paper submission taxonomy, e.g., "Computer Vision" and "Game Theory". Each part of the taxonomy is looked at from the perspective of "Research Question", "XAI Challenge",  "Methods", "Limitations", and "Opportunity". Their taxonomy gives researchers an excellent introduction to how CAI on KGs can be leveraged in different fields of AI, what the pitfalls are, and where future research directions lie. In contrast, the survey at hand aims to provide a complete overview of existing CAI methods on KGs and to cluster them in the taxonomy proposed in Section \ref{sub:scop}. 
Three other related surveys were identified. However, they focus on comprehensible recommender systems \citep{Zhang2020b}, comprehensible Graph Neural Networks \citep{Yuan2020}, and comprehensible semantic web technologies \citep{Seeliger2019}, including non-KG-related ML methods. 

The following section shall provide relevant theoretical background for the concepts that form our taxonomy (cf. Table \ref{tab:scop}).
\section{Theoretical Background and Taxonomy of Comprehensible Artificial Intelligence on Knowledge Graphs}\label{sub:meth}
CAI, XAI as well as IML are sometimes used interchangeably \citep{Zhang2020, Zhang2021, Wang2018}. We trace this back to the fact that especially the concept XAI is overloaded and, at the same time, not sharply differentiated from CAI and IML. Therefore, this section shall provide a structured, theoretical background on these notions and outline the concepts of our taxonomy in detail. \\
In previous literature, XAI describes inherently interpretable AI models like \emph{TEM} \citep{Wang2018} (cf. Table~\ref{tab:literature_table_iml}) and post-hoc explanation methods like \emph{RelEx} \citep{Zhang2021} (cf. Table~\ref{tab:literature_table_xai}). 
This includes post-hoc explanation methods that give third-party rationalization of the AI model's decision \citep{LULLY2018211, Tiddi2021}. 
Meanwhile, there is the concept IML. IML models are  inherently interpretable \citep{Rudin2019}, which makes the concept overlap with a particular usage of XAI. The fuzzy border and the overload are due to similar high-level needs satisfied by IML and XAI, like identifying risks of the AI model's output or providing an ethical justification of the model's output \citep{Adadi2018}. However, inherently interpretable AI models and post-hoc explanations methods are fundamentally different. One is a complete ML model able to classify its input, and the other is a technique unable to classify on its own. Thus, both should be described with different concepts. 
This survey accounts for the fundamental differences in the inherently interpretable AI models and post-hoc explanation methods by grouping them in IML and XAI. A clear-cut definition for both concepts shall be provided in this section. In addition, this survey accounts for the similar high-level needs IML and XAI satisfy by introducing CAI as their parent concept in the papers' taxonomy. 

This survey follows the understanding of the concepts XAI and IML as introduced in work by \citet{Rudin2019} and \citet{Schmid2021a}. Composing the concept of XAI and IML to the overarching concept of CAI is also part of the contribution of this writing. The concepts CAI, XAI, and IML shall be outlined in the following sections. 

\subsection{Comprehensibility}\label{sub:comp}
CAI is about opening up black box models and explaining input, inside reasoning and processing, and output in an adequate manner to the relevant users. 
CAI makes a black box AI model transparent without decreasing its performance. 
Another side of CAI is developing algorithms that produce interpretable ML models, so-called white-box models. 
Those white-box models are supposed to perform close to the state-of-the-art, in order to reach widespread application \citep{Adadi2018}.

CAI serves the following four reasons. 

(1) The CAI model is supposed to justify its output \citep{Adadi2018, Lipton2018}. One possible application of this is to verify that the resulting output aligns with the ethical principles of a particular moral framework \citep{Adadi2018, Lipton2018}.

(2) Furthermore, explanations help to identify risks and flaws of an AI model.
A possible risk is the \emph{Clever Hans} bias described by \citet{Lapuschkin2019}. They describe an image classifier that classifies horses correctly because all pictures with horses have a particular watermark on them. The explanation of the classifier uncovers this flaw \citep{Adadi2018, Lipton2018}. 

(3) Finding risks and flaws of an AI model enable researchers or developers to debug and improve the model \citep{Adadi2018, Lipton2018}.

(4) In addition, producing explanations for an AI model might uncover unknown rules discovered by the AI model \citep{Adadi2018, Lipton2018}. 



Figure~\ref{fig:cai} displays the generic CAI framework separating XAI from IML. The following shall elaborate more on the notions XAI and IML, contributing to a clear-cut notion of CAI.
\subsubsection{Interpretability}
\label{subsubsec:iml}
IML methods require a
per default human-understandable decision-making process
to integrate the user into the system, rather than offering a set of descriptive tools \citep{Lahav2018, molnarInterpretableMachineLearning2019, Rudin2019, Schmid2021a}. Thus, a IML model is able to classify and to provide a meaningful explanation for a classification by itself. The explanation is faithful to the model's decision-making process and behaviour. Such methods, shall, for the scope of this survey, be referred to as IML systems. 
XAI will be outlined in the following.

\subsubsection{Explainability}
While IML tries to uncover the whole decision-making process of the ML model, XAI is about mapping the input of a black box model to its output. 
That way, XAI methods compute an explanation of the ML model’s behaviour. As the explanation method is external to the ML model, faithfulness is not guaranteed. However, the quality of a XAI method is measured by how faithful their explanations reflect the model's behaviour. The inside reasoning and processing of the black box model are not be made transparent by the XAI method.

A common theme in XAI is the attribution of a metric that represents the impact of a specific input characterization, frequently referred to as the attribution value of an input feature. The attribution tells how important an input feature is for or against a classification. Such methods are so-called attribution methods \citep{Zhou2021, Zhai2006, Ribeiro2016, Lundberg2017}. 
Other innovative themes exist in XAI, like counterfactual explanation methods \citep{Verma2020} or explanations methods based on Inductive Logic Programming (ILP) \citep{Rabold2018}. \\

CAI significantly benefits from semantic information. KGs provide semantic information. Thus, the following sections shall introduce the KGs and the taxonomy of CAI on KGs.

\subsection{Knowledge Graphs}\label{sub:kg}
\begin{figure}[h!]
    \centering
    \includegraphics[width=0.6\linewidth]{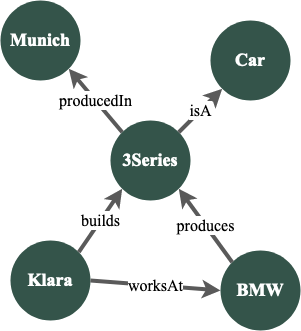}
    \caption{Car manufacturer KG.}
    \label{fig:running_example_kg}
\end{figure}

A KG is a directed labeled graph \citep{Hogan_2022}  
(cf. Figure \ref{fig:running_example_kg}).

    In general, a graph is a set of connected entities. 
    Entities are referred to as vertices $v$ (or nodes), e.g., ``\emph{Klara}", while connectors are referred to as edges $e$ (or relations), e.g., ``\emph{builds}", in the context of a graph.
    Thus, a graph $U$ consists of a set of vertices $V = \{v_1,..., v_i, ..., v_n\}$ and edges $E = \{e_1, ..., e_j, ..., e_m\}$.  This renders a graph $U$ as a subset of $V \times E \times V$ \citep{sanchez-lengelingGentleIntroductionGraph2021}.

    A directed graph is a graph in which edges have a direction. Thus, given a directed edge $e_1$ from vertex $v_1$ to vertex $v_2$, 
graph traversal via $e_1$ is exclusively possible from $v_1$ to $v_2$ and not vice versa \citep{Harary1965}. A labeled graph is a graph in which holds $ \forall v_i \in V$ and $ \forall e_j \in E$ that exactly one element from a set of symbols $S = \{s_1, ..., s_k, ..., s_o\}$ is assigned to them. A symbol $s_k$ is, in general, the name of an entity like ``\emph{Klara}"  or the name of a relationship like ``\emph{worksAt}" \citep{Gallian2001}. A head entity $h \in V$ connected via a relation $r \in E$ to a tail entity $t \in V$, e.g., ``\emph{Klara worksAt BMW}", is called a triple. Due to the latter properties, a KG is optimal for storing highly relational data while preserving its semantics. 

\subsection{A taxonomy of Comprehensible Artificial Intelligence on Knowledge Graphs}

The section introduces the critical aspects of CAI on KGs. First, the representations of KGs shall be discussed, followed by the tasks, and foundations of CAI on KGs.

\subsubsection{Representations}\label{sub:represenation}
KGs as inputs to AI models are represented sub-symbolically, symbolically, or neuro-symbolically \citep{Hogan_2022}. 

\begin{figure}[h]
\begin{minipage}{0.25\textwidth}
\begin{tikzpicture}[scale=0.5]
\tikzstyle{every node}=[font=\LARGE]
\begin{axis}[
  axis line style={Stealth-Stealth},
  xmin=-0.5,xmax=3.,ymin=-1.,ymax=1.5,
  xtick distance=1,
  ytick distance=1,
  xlabel=$x$,
  ylabel=$y$,
  grid=major,
  grid style={thin,densely dotted,black!20}]
\node[label={left:{BMW}},circle,fill,inner sep=2.5pt] at (axis cs:2.5,0.5) {};
\node[label={right:{Klara}},circle,fill,inner sep=2.5pt] at (axis cs:0,1) {};
\coordinate (S) at (axis cs:0,0);
\coordinate (C) at (axis cs:2.5, -0.5);
\draw[,very thick,->](S)--(C) node [midway, fill=white] {worksAt};
\end{axis}
\end{tikzpicture} 
\end{minipage}%
\begin{minipage}{.25\textwidth}
\begin{equation*} 
\begin{split}
    Klara = (0.0, 1.0) \\
    BMW = (2.5, 0.5) \\
    \vec{worksAt} = (2.5, -0.5)
\end{split}
\end{equation*}
\end{minipage}
\caption{Representation of the entities and relation, \emph{Klara}, \emph{BMW}, and \emph{worksAt} in $\mathbb{R}^2$ ($x, y \in \mathbb{R} $).}\label{pic:transE}
\end{figure}
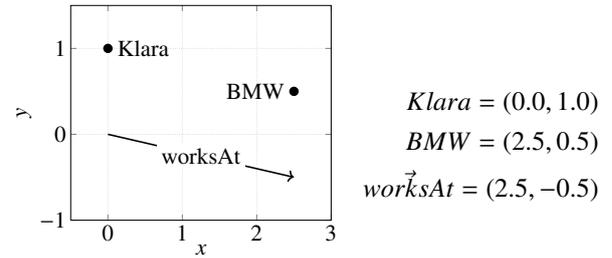

The sub-symbolic representation of KGs gained vast popularity with the rise of scalable neural ML. The general idea is to embed the entities and relations of the KG into low dimensional numeric space, like an adjacency matrix or a real-valued matrix \citep{Bordes2013}. An example of entities and relations in $\mathbb{R}^2$ is depicted in Figure \ref{pic:transE}. However, at scale, such representations become incomprehensible for humans. 

A KG is by default a symbolic structure, as the nodes and edges of a KG are labelled with symbols. Thus, unlike the sub-symbolic representation, no specific mapping is required to achieve a symbolic representation of a KG. 
The symbolic representation is human comprehensible and allows for reasoning, inconsistency checking, and various other symbolic AI methods (cf. Section \ref{subsec:iml_on_kgs}).   

Combining sub-symbolic and symbolic representations, while leveraging the advantages of both representations, achieves a neuro-symbolic\footnote{Neuro-symbolic AI is also known as the third wave of AI \citep{Schmid2021b, Garcez2020}.}
 representation \citep{Amato2022} (cf. Section \ref{subsec:iml_on_kgs}).



\subsubsection{Tasks} \label{sub:tasks}
We differentiate five different tasks of CAI on KGs, namely Link Prediction (LP), Node Clustering (NC), Graph Clustering (GC) and Recommendation (R).

Clustering can be divided into vertices-clustering \citep{Xiao_2021} and structural clustering \citep{Huang2010LinkPB,Tian2014LearningDR}. Vertices-clustering clusters vertices by their similarity, like the bi-section k-modes clustering \citep{Schmitz2006}. In contrast, structural clustering uses the graph topology as input to cluster the whole graph, like the \emph{Louvain} clustering algorithm \citep{Elbattah2017}. 

The recommendation task is about finding a set of relevant items for a user \citep{Liu2021}. Users and items may be modeled as entities of the KG.  
\citet{Liu2021} provide a comprehensive survey on KG-based recommender systems. 


KGs are notoriously incomplete, i.e, relations between two entities might be missing. Predicting such missing relations, e.g., ``\emph{Klara worksIn Munich}", is called link prediction \citep{Bordes2013}. 

To classify a vertice in the KG regarding a set of classes external to the KG 
is called vertice classification \citep{Steenwinckel2022}, e.g., ``\emph{Klara}" is an ``\emph{Employee}". 

Furthermore, it might be interesting to classify the KG as a whole. Molecules, for example, can be represented as KGs \citep{hwang2020comprehensive}. Graph classification is about classifying whether the molecule is toxic \citep{Lee2018}. 

The following section shall give short examples of how foundations for those tasks are designed.  

\subsubsection{Foundations} \label{sub:optimization}
There are numerous and diverse foundational ML methods for CAI on KGs. Only a few of them can be discussed due to the scope of the paper.  
Translational öearning, Neural Networks, in particular Graph Neural Networks, and rule-based learning are some of the most influential foundation (cf. Section \ref{sub:related_work}), which are introduced in the following.

Translational learning methods translate the entities and relations of a KG into numeric space. This is achieved by learning a score function. The score function expresses a predefined property of the translational method \citep{Ali2020}. For example, the score function of \emph{TransE} is $d(h+r,t)=||h+r-t||$ \citep{Bordes2013}. The score function is optimized by a loss function to embed every element of a triple $(h,r,t) \in \mathbb{R}^2$, such that the relation added to a head entity results in the representation of the tail entity (cf. Figure \ref{pic:transE})\citep{Bordes2013}.

Popular connectionism methods are Neural Networks and Graph Neural Networks in the context of graphs.
Graph Neural Networks are non-linear functions in the form of layered networks that can be optimized by gradient descent on all subsets of the graph $U$ to perform regression or to learn classes or concepts \citep{scarselliGraphNeuralNetwork2009}. 
However, standard Graph Neural Networks cannot integrate labels and directions of relations \citep{scarselliGraphNeuralNetwork2009}. 

Rule-based learning methods heuristically search through possible combinations of logical clauses \citep{Cohen1995,Svatek2006,Lao2011}. The resulting clauses, i.e. rules, optimize the coverage of triples, predictive accuracy, and other task-specific goal functions \citep{Galarraga2015}. 

Other foundations exist, like Factorization Machines (FM) and Reinforcement Learning (RL). The authors refer to \citet{bokde2015matrix} and \citet{sutton2018reinforcement} for a comprehensive introduction to those methods.

The components by which we taxonomize CAI on KGs are completely presented. In the following section, specific methods shall be discussed in accordance with our taxonomy. 
\section{Comprehensible Artificial Intelligence on Knowledge Graphs}\label{sec:kg_for_cai}
\begin{figure}[ht]
    \centering
    \includegraphics[scale=0.26]{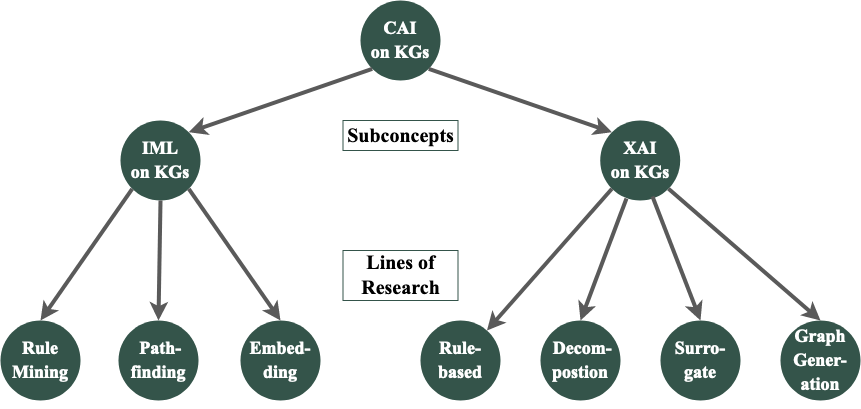}
    \caption{Hierarchy of the lines of research in CAI on KGs.}\label{fig:cai_on_kgs_lines_of_research}
\end{figure}

We identified different lines of research in IML and XAI on KGs during the literature review (cf. Figure~\ref{fig:cai_on_kgs_lines_of_research}).
There are three lines of research in IML on KGs, which are 
\begin{enumerate}
\item Rule mining methods, 
\item Pathfinding methods, and
\item Embedding methods, 
\end{enumerate}
and four lines of research in XAI on KGs, which are 
\begin{enumerate}
\item Rule-based learning methods, 
\item Decomposition-based methods, 
\item Surrogate methods, and 
\item Graph generation methods.
\end{enumerate}
IML and XAI on KGs are fundamentally different but similar in the high-level needs they satisfy. The fundamental difference leads to unique lines of research in both concepts, and a varying number in lines of research this survey identified for both concepts.

The following subsections shall introduce the lines of research in IML and XAI on KGs. The lines of research are described by presenting relevant. The aim is to draw a picture of how the research lines evolved and where past and current challenges lie. 

\subsection{Interpretable Machine Learning on Knowledge Graphs} \label{subsec:iml_on_kgs}
IML methods on KGs come in many shapes and colors. As argued in \ref{subsubsec:iml}, the main commonality of such models is that they are self-explanatory white-box models for prediction and classification tasks on KGs. This survey identifies several lines of research in the field of IML on KGs. 
One of the earliest lines of research are rule mining methods.
\begin{table*}[h!]
  \centering
  \begin{tabular}{p{3cm}| p{4cm}| p{5.1cm}}
    \hline
     \multicolumn{3}{c}{Model Interpretaion via}  \\
     \hline
    Clause & Path & Attention Vector \\ 
    \hline
    \begin{minipage}{32mm}
    \scalebox{0.8}{
        $builds(x,y)$ 
        }
    \scalebox{0.8}{
        $\land \, producedIn(y,z)$
        }
    \scalebox{0.8}{
        $\implies worksIn(x,z) $
        }
    \end{minipage}
    & 
    \begin{minipage}{40mm}
    \includegraphics[width=\linewidth]{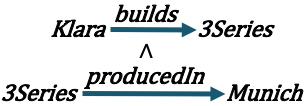} 
    \end{minipage}
    &
    \begin{minipage}{51mm}
    \includegraphics[width=\linewidth]{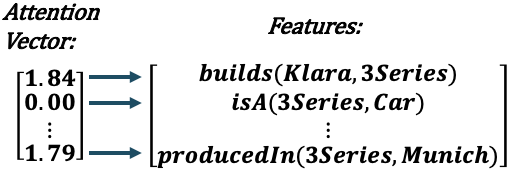} 
    \end{minipage}
  \end{tabular}
  \caption{Three different types of explanations generated by interpretable models for the link prediction task $"Klara \  worksIn \   ??$" with the answer "$Munich$".}\label{tbl:example_explanations}
\end{table*}
\subsubsection{Rule mining methods}
One may trace back the roots of rule mining methods on KGs to ILP \citep{Muggleton1991} algorithms like \emph{FOIL} \citep{Quinlan1990}. However, their scope is learning rules from any logical knowledge base. Thus, traditional ILP algorithms are not optimized to scale to large KGs \citep{Cohen1995,Svatek2006,Lao2011}. Rule mining methods customized to IML on KGs improve the scalability.

One of the first and most impactful white-box rule mining algorithms specifically designed for KGs was \emph{PRA} \citep{Lao2011}. \emph{PRA} samples numerous path types. Next, random walks through the KG are conducted. The random walks are constrained to follow one of the previously sampled path types. The random walks count the frequency of vertex-pairs being the start and end of the same path type. Finally, the frequencies of possible path types over the vertices are matched via logistic regression. That way, general rules, i.e. interpretable clauses (cf. Table \ref{tbl:example_explanations}), are inferred. The clauses may be used to predict missing links between vertices. 
Furthermore, the path type frequencies might be used as the embedding of a vertex \citep{Lao2011}.
While \emph{PRA} has increased scalability, random walks are still computationally expensive. In addition, the link prediction accuracy is not state-of-the-art since factorization and translation methods, like \emph{RESCAL} \citep{Nickel2011} and \emph{TransE} \citep{Bordes2013}, were introduced. Research following \emph{PRA}, like \emph{SWARM} \citep{Barati2016}, \emph{RLvLR} \citep{Omran2018}, \emph{RDF2rules} \citep{Wang2015}, and \emph{AIME+} \citep{Galarraga2015} on rule-based algorithms addresses those issues. In particular, \emph{AIME+} is another ILP-inspired algorithm for scalable rule mining and link prediction on KGs. \emph{AIME+} builds rules for link prediction by iteratively adding new clauses to a rule such that the approximated coverage and confidence of the new rule is maximized \citep{Galarraga2015}. 
\emph{AIME+} improved on scalability and predictive performance compared to \emph{PRA}. 
Ontology Pathfinding (\emph{OP}) \citep{Chen2016a}, and \emph{ScaLeKB} \citep{Chen2016b}, scales the idea of \emph{AIME+} to  large benchmark KGs datasets like \emph{Freebase}. It does so by parallelizing join queries, breaking the rule mining task into smaller sub-tasks, and eliminating unsound and resource-consuming rules before extending them.
In addition, \emph{AnyBURL} \citep{Meilicke2019AnytimeBR, Meilicke2020ReinforcedAB} and subsequently \emph{SAFRAN} \citep{Ott2020} build on \emph{PRA} to increase its scalability by an anytime bottom-up technique for rule learning \citep{Meilicke2019AnytimeBR} and aggregating rules through a scalable clustering algorithm \citep{Ott2020}. These techniques allow rule mining methods, like \emph{AnyBURL} and \emph{SAFRAN}, to scale to modern large-scale KGs while showing state-of-the-art predictive performance \citep{rossi2021}.  

One significant research effort was to scale symbolic rule mining approaches to large KGs, which was subsequently achieved by highly optimized heuristics and rule learning techniques. The following methods take a different path to scale rule mining to large KGs. They combine symbolic and sub-symbolic representations of the KG to efficiently mine interpretable rules.
\emph{NeuralLP} \citep{Yang2017} mines interpretable rules from KGs. \emph{NeuralLP} is inspired by the probabilistic deductive database \emph{TensorLog} \citep{Cohen2017}. \emph{NeuralLP} learns weighted chain-like logical rules. Those link the head of a query to its tail. It does so by embedding relations in an end-to-end differentiable model as operators. A recurrent neural network learns how to decompose these operators, such that the loss of the link prediction task is optimized \citep{Yang2017}. 
\emph{DRUM} \citep{Sadeghian2019} builds on the idea of \emph{NeuralLP}. It increases the expressivity of \emph{NeuralLP} by an improved representation of the differentiable operators. This allows for learning rules that cover a higher number of relations than \emph{NeuralLP}. Furthermore, it uses a bidirectional recurrent neural network to optimize the operators. The bidirectional network capture's information about the backward and forward order in which the relations can appear in the rule \citep{Sadeghian2019}. \citet{Sadeghian2019} show that \emph{DRUM} outperforms state-of-the-art link prediction models like \emph{MINERVA} \citep{Das2017} and \emph{RotatE} \citep{Sun2019} on the inductive link prediction task. 
\emph{RuLES} \citep{Ho2018} is also a neuro-symbolic method to rule mining. It learns clauses guided by a KG embedding. At each iteration, \emph{RuLES} expands the existing clause such that the link prediction performance is optimized. The KG embedding provides the probability of a clause to be expanded by an atom \citep{Ho2018}. That way, the search space for new clauses is effectively traversed, which increases the scalability and performance of \emph{RuLES}. 
\citet{Ma2019} induce explainable rules from KGs by creating a joint learning framework. The authors construct a rule-guided neural recommendation model, whereby inductive rules are mined from item-centric KGs. Other neuro-symbolic methods mine rules with the help of reinforcement learning \citep{CHEN2022108371}. First, an RL agent builds clauses with a curriculum learning strategy. Next, a rule mining system is built by applying the value function of the RL agent to efficiently guide the search for high-quality rules \citep{CHEN2022108371}.

\subsubsection{Pathfinding methods} 
The latter methods mine interpretable rules from KGs, which then are used to predict missing links between entities. This method is straightforward in discovering knowledge and building white-box AI models to predict missing links in the KG. Another line of research trains RL agents to find a path through KGs, the walked paths are the explanation of the prediction. A path-based interpretation is depicted in Table \ref{tbl:example_explanations}.

\emph{DeepPath} \citep{Xiong2017} builds the bedrock of those methods. It uses \emph{TransE} \citep{Bordes2013} to predict the tail for a head-relationship pair. Next, \citet{Xiong2017} train a RL agent to find the most informative path linking the head and tail entity of the predicted triple. However, as the pathfinding is decoupled from the actual link prediction process, the faithfulness required for an explanation method is not given in the case of \emph{DeepPath} \citep{Xiong2017}. 
The method inspired research on interpretable link prediction from the perspective of a pathfinding problem.

The first interpretable method deploying this perspective was \emph{MINERVA} \citep{Das2017, Das2018}. \emph{MINERVA} treats the KG vertices as states, and relations are the actions or paths to transition from one state to the other. A RL agent is trained to walk the optimal path through the KG to the correct tail entity. The policy network of the RL agent is an \emph{LSTM} \citep{Hochreiter1997}. This allows the agent to consider the already walked path when choosing an action. The agent is rewarded whenever it walks to the correct tail entity. If the agent gets lost, it is not rewarded \citep{Das2018}. 
\citet{Lin2018} observe that in an incomplete KG the RL agent receives low-quality rewards by false negatives in the training data. This reduces the generalization capacity of the agent at inference. In addition, \emph{MINERVA} provides no optimal paths to the agent at training time. Thus, paths can be learned that accidentally lead to the correct answer. They address this problem by introducing a pre-trained one-hop embedding model, like \emph{ComplEx} \citep{Trouillon2016} or \emph{ConvE} \citep{Dettmers2018} to estimate the reward of unobserved facts. \citet{Lin2018} show that this reduces the impact of false negatives in the training data.
Furthermore, \citet{Lin2018} force the agent to explore a diverse set of paths by applying randomly generated edge masks to the KG environment at training time. 
\citet{Hou2021} are motivated by the same issues addressed by \citet{Lin2018}. They adapt the search strategy of the RL agent. While in training, the agent selects paths guided by high-quality rules from a rule pool \citep{Hou2021}. This results in less low-quality rewards and mitigates the impact of spurious paths.
\citet{Bhowmik2020} extend the path-finding method to the inductive link prediction domain. This allows link prediction for previously unseen entities. They achieve this by using a graph attention network for policy selection of the RL agent and a pre-trained \emph{ConvE} \citep{Dettmers2018} embedding as a reward estimator \citep{Bhowmik2020}.
Furthermore, the interpretable pathfinding method on KGs is employed in various recommender systems 
\citep{Xian2019, Song2019ExplainableKG, Zhang2020}. 
In particular, \emph{LOGER} \citep{Zhu2021} deploys a \emph{Markov Logic Network} \citep{Qu2019} to guide the RL agent to the optimal result. The \emph{Markov Logic Network} learns personalized importance scores over paths from the user to recommendation items. To do so, the \emph{Markov Logic Network} learns from previous user behaviour. Next, a RL agent finds the optimal path from the user to the recommendation item by optimizing the correctness of the recommendation and the importance scores of a path \citep{Zhu2021}. 

\subsubsection{Embedding methods}
Standard embedding methods show high predictive performance. However, they are not interpretable. A recent line of research tries to mitigate this issue by designing interpretable embedding models. 

\citet{Xie2017} proposes \emph{ITransF}. 
\emph{ITransF} consists of a translational embedding method similar to \emph{STransE} \citep{Nguyen2016}. In addition, they learn sparse attention vectors. The sparse attention vectors reflect how important a concept dimension is for the classification, and thus help to understand which underlying concepts different relations share \citep{Xie2017}. An attention-based interpretation is depicted in Table \ref{tbl:example_explanations}. Attention vectors are not the only way the literature explores to achieve interpretability in sub-symbolic methods.
\citet{anelliHowMakeLatent2019} introduces an interpretable recommender method relying on an FM. The latent factors of the FM are initialized by semantic features coming from a KG \citep{anelliHowMakeLatent2019}. In addition, semantic features are injected into the learning process. \citet{anelliHowMakeLatent2019} show that this guarantees the interpretability of the methode's recommendations. 
\emph{CrossE} \citep{Zhang2019a} is a sub-symbolic embedding for link prediction that explains by analogy. \emph{CrossE} predicts missing links by a translational model. Next, \emph{CrossE} generates explanations of the predicted link. It does so by finding closed paths from the head to the predicted tail entity. Next, it searches for similar structured paths in the KG between entities already connected by the predicted relationship \citep{Zhang2019a}. The found paths are shown to the user to justify the models' predicted link \citep{Zhang2019a}. 
Another embedding-based method is called \emph{INK} \citep{Steenwinckel2022}. \emph{INK} is used to create embeddings of vertices given their neighborhood. The embeddings may be used in downstream ML models for vertex-level classification \citep{Steenwinckel2022}. \emph{INK} learns a binary feature representation of the vertex's neighborhood for every vertex. The binary features allow users to interpret and understand the embeddings create by \emph{INK} \citep{Steenwinckel2022}. \citet{Steenwinckel2022} show that \emph{INK} performance is similar to the state-of-the-art black box methods, like \emph{R-GCN} \citep{Schlichtkrull2018} and \emph{RDF2Vec} \citep{Ristoski2016}.

\emph{TEM} is neuro-symbolic method, relying on the attention vectors to establish interpretability \citep{Wang2018}. First, \emph{TEM} uses a decision tree to learn decision rules. Next, they design an attention network-based embedding model that uses the decision rules as features. The combination of the decision tree and the attention network guarantees that the recommendation process is fully interpretable \citep{Wang2018}.
\citet{Ai_2018} propose a different neuro-symbolic embedding method. The authors use at the core of their method a translation function similar to \emph{TransE} \citep{Bordes2013}. 
The embedding is 
used to recommend items \citep{Ai_2018}. Next, the model generates an explanation \citep{Ai_2018}. First, possible closed paths between the head and tail entity are generated by breadth-first search. Then, the probability for each path is calculated by the translational function. The path with the highest probability is shown to a user to explain the predicted link \citep{Ai_2018}. Notably, 
\citet{Ruschel2019} propose a similar method. However, they focus on the link prediction task instead of the recommendation task and use \emph{SFE} \citep{Gardner2015, Gusmao2018}, a performance version of \emph{PRA} \citep{Lao2011}, to mine rules used as explanations. 
\citet{Polleti2020} also combine a rule mining and embedding method to compute interpretable recommendations. However, they add counterfactuals to their explanation. Thus, they do not simply justify why the user should trust the recommendation, but also why the user should not trust it. They do so by implementing \emph{Snedegar's} theory of reasoning. For example, they select reasons, i.e. paths through the KG, against a recommendation. They do so by searching for reasons in favor of a competing recommendation that the initial recommendation is not satisfying \citep{Polleti2020}. 

Titles related to and/or discussed in this section are grouped regarding the survey's taxonomy in Table \ref{tab:literature_table_iml}. The following section completes the taxonomy in regard to XAI on KGs.

\begin{table*}[!ht]
\centering
    \begin{tabular}{ | p{5.25cm} | p{4.25cm} | p{2cm}  p{1cm}  p{1.75cm} |}
    \toprule
    {\textbf{Line of research}} & {\textbf{References}}& {\textbf{Representation}} & {\textbf{Task}} & {\textbf{Foundation}} \\
    \toprule
    \textbf{Rule Mining Methods}    &   &   &   &  \\
    \multirow{8}{*}{\small{\makecell[l]{Mining human-understandable clauses \\
    by optimizing hypothesis space travesel \\ via effective heuristics \\ and efficient search strategies.}}} & \emph{PRA} \citep{Lao2011}      & \multirow{8}{*}{ \emph{SB}} &  \multirow{8}{*}{LP} & \multirow{8}{*}{RBL} \\
    & \emph{SWARM} \citep{Barati2016}   &   &   &     \\
    & \emph{RLvLR} \citep{Omran2018}    &   &   &     \\
    & \emph{RDF2Rules} \citep{Wang2015} &   &   &     \\
    & \emph{AIME+} \citep{Galarraga2015} &  &   &     \\
    & \emph{OP} \citep{Chen2016a} &  & &    \\
    & \emph{ScaLeKB} \citep{Chen2016b} &  & & \\
    & \emph{AnyBURL} \citep{Meilicke2019AnytimeBR, Meilicke2020ReinforcedAB} &  &   &   \\
    & \emph{SAFRAN} \citep{Ott2020} &   &   & \\                                              
    \hline
    \multirow{2}{*}{\small{\makecell[l]{Mining human-understandable clauses \\ with the help of differentiable operators.}}}   & \emph{NeuralLP} \citep{Yang2017} & \multirow{4}{*}{\emph{NSB}} & \multirow{4}{*}{LP} & \multirow{4}{*}{RBL} \\
    & \emph{DRUM} \citep{Sadeghian2019}    &   &   & \\
    \cline{1-2}
    \small{\makecell[l]{Mining human-understandable clauses \\ guided by a KG embedding.}} & \emph{RuLES} \citep{Ho2018}    &  &  &    \\
    \hline
    \small{\makecell[l]{Mining human-understandable clauses \\ via rule-guided neural recommendation \\ model.}} & \emph{RULEREC} \citep{Ma2019} & \emph{NSB} & R & RBL  \\ 
    \hline
    \small{\makecell[l]{Mining human-understandable clauses \\ by guiding the rule search via a RL agent.}} & Reinforce Rule Miner \citep{CHEN2022108371} & \emph{NSB} & LP & RL \\
    \toprule
    \textbf{Pathfinding Methods} & & & & \\
    \multirow{8}{*}{\small{\makecell[l]{Finding human-understandable paths \\ through the knowledge graph.}}} & \emph{MINERVA} \citep{Das2017} & \multirow{4}{*}{ \emph{NSB}} & \multirow{4}{*}{LP} & \multirow{4}{*}{RL} \\
    & \emph{Multi-hop} \citep{Lin2018} & & & \\
    & Inductive Pathfinding \citep{Bhowmik2020} & & & \\
    & \emph{RARL} \citep{Hou2021} & & & \\
    \cline{2-5}
    & \emph{PGPR} \citep{Xian2019} & \multirow{4}{*}{ \emph{NSB}} & \multirow{4}{*}{R} & \multirow{4}{*}{RL} \\
    & \emph{EKAR} \citep{Song2019ExplainableKG} &  &  & \\
    & Distiling Knowledge \citep{Zhang2020}  & & & \\
    & \emph{LOGER} \citep{Zhu2021} & & & \\
    \toprule
    \textbf{Embedding Methods} & & & & \\
    \small{\makecell[l]{Generating semantic features \\ from the KG as interpretations.}} & Feeding KG \citep{anelliHowMakeLatent2019} & \emph{SSB} & R & FM \\
    \hline
    \small{\makecell[l]{Generating attention mask \\ as interpretations.}}  & \emph{ITransF} \citep{Xie2017}, & \multirow{3}{*}{\emph{SSB}}   & \multirow{3}{*}{LP} & \multirow{3}{*}{TL} \\
    \cline{1-2}
    \small{\makecell[l]{Interpreting LP by analogies \\ found in the KG.}} & \emph{CrossE} \citep{Zhang2019a}  &  & & \\
    \hline                                                        
    \small{\makecell[l]{Interpretability via Decision Tree \\ and Attention network.}} & \emph{TEM} \citep{Wang2018} & \emph{NSB} & R & \emph{O} \\ 
    \hline
    \small{\makecell[l]{Interpretable binary \\ neighborhood representations.}} & \emph{INK} \citep{Steenwinckel2022} & \emph{SSB} & NC & TL   \\
    \hline
    \small{\makecell[l]{Generating and selecting rules \\ via SFE and embedding function.}} & \emph{XKE-e} \citep{Ruschel2019} & \emph{NSB} & LP & TL  \\
    \hline
    \small{\makecell[l]{Generating and selecting rules \\ via BFS and embedding function.}} &  Heterogeneous KGE \citep{Ai_2018} & \multirow{3}{*}{\emph{NSB}}   & \multirow{3}{*}{R} & \multirow{3}{*}{TL} \\
    \cline{1-2}
    \small{\makecell[l]{Generating reasons for and \\ against a recommendation.}} & Snedegar Reasoning \citep{Polleti2020} & &  & \\
    \bottomrule
    \end{tabular}
    \normalsize
    \caption{Literature review table of IML on KGs. Each item is classified by representation, task, foundational method, and line of research. Sometimes, approaches are constructed by multiple publications building upon each other. In this case, all related publications are cited.}\label{tab:literature_table_iml}
\end{table*}

\subsection{Explainable Artificial Intelligence on Knowledge Graphs}\label{sub:kgforxai}
Recalling that XAI is the mapping of the output of a ML model to its corresponding input (cf. section~\ref{sub:comp}), this section provides an overview of those methods. Again, the following sections are structured based on the lines of research (cf. Figure~\ref{fig:cai_on_kgs_lines_of_research}) from our taxonomy for XAI on KGs.
\subsubsection{Rule-based learning methods}
Typically, RBL applications identify patterns by rules induced from the underlying data. 
\citet{Donadello2021}, create the framework \emph{SeXAI}, where human-understandable First Order Logic (FOL) formulas are produced. These formulas consist of features connected to the classes of a knowledge base. Furthermore, their connection to both the black box model and the annotations of the dataset allows for the explanation of the neural network predictions via semantic features. This renders the method neural-symbolic.\footnote{FOL quantifies individual objects as variables, while higher-order logic also quantifies sets of individual objects.} Other FOL based methods exist, such as \citet{Dervakos2022}, who uses the symbolic representation of a KG to express FOL rules in the terminology of the KG.
Furthermore, \emph{ExCut} \citep{Gad-Elrad2020} lives on the edge between RBL and GC. \citet{Gad-Elrad2020} address the issue of latent embeddings. The authors show that latent embeddings have no significant contribution to the explainability of a prediction. Furthermore, their method creates human-understandable labels for clusters by applying rule mining methods on KG embeddings. The learned rules are used to generate labels for clusters, formed by an graph clustering method. The clustering method is embedding-based and clusters entities of a graph and is another contribution of the paper. Eventually, those labels are formed by combining entity relations according to the so-called cluster explanation rules.
\subsubsection{Decomposition methods}
\begin{figure*}[!ht]
  \centering
  \includegraphics[width=\textwidth]{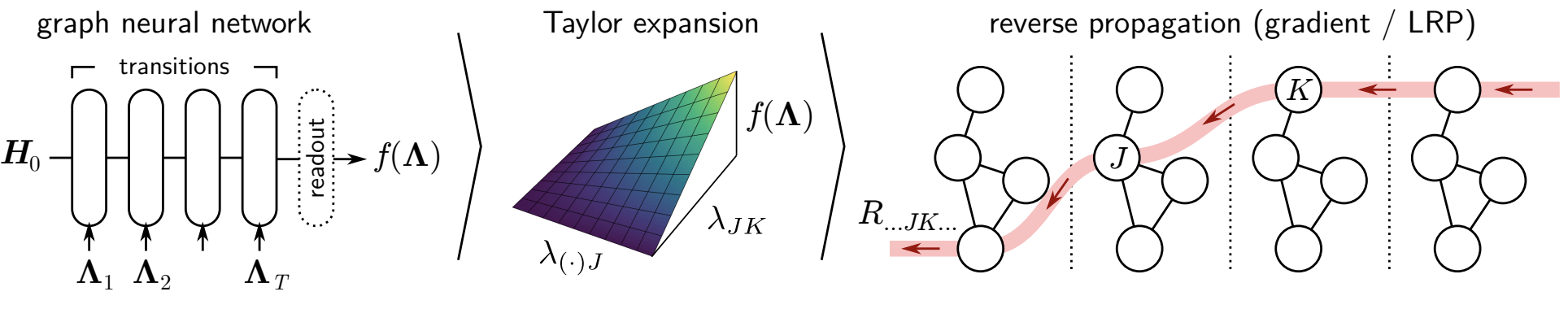}
  \caption{Exemplary GNN LRP method, taken from \citep{Montavon2019}.}\label{fig:lrp}
\end{figure*}

A popular method to measure the importance of input features of a black box model is to decompose its predictions into separate terms. The partition of the prediction score of each term is then interpreted as the importance measure of the respective input feature.
Recently, decomposition methods on Graph Neural Networks were proposed by \citet{Schnake2020}, who elaborated on high-order explanations of Graph Neural Networks via so-called relevant walks. The authors managed to identify groups of edges that commonly contribute to the GNN prediction by using a nested attribution scheme. At each step of this scheme, layer-wise relevance propagation (LRP) 
\citep{Bach2015, Montavon2019}
is applied in order to determine relevance. In accordance with our definition of XAI, the GNN LRP method explains any GNN prediction by extracting the most relevant paths from the input to the output of the GNN. Here, paths correspond to walks on the input graph $U$. The method is based on high-order \emph{Taylor} expansions of the GNN output as well as chain-rules that are used to extract the \emph{Taylor} expansion's terms \citep{Bach2015}. Eventually, the GNN LRP method is transferable to large, nonlinear models. The method is applicable to a vast range of Graph Neural Network architectures and is set for application in sentiment analysis of text data, structure-property relationships in quantum chemistry, and image classification \citep{Montavon2019}. LRP remains one of the most popular ANN explanation methods, however, a detailed description of the GNN LRP method would exceed the scope of this survey. Figure~\ref{fig:lrp} summarizes the proceeding.
Furthermore, \citet{NEURIPS2019_d80b7040} introduce \emph{GNNExplainer}, which highlights those modules of the GNN that explain the prediction. The authors determine these modules by learning the most suitable explanations via graph perturbation. The main idea is to perturb the input graph by adding or removing edges, nodes, or features in order to observe how these changes affect the Graph Neural Networks output. This perturbation involves making small changes to the graph's nodes and edges while keeping the rest of the structure intact. The technique is based on the assumption that the gradient of the prediction is known and the underlying data is independent and identically distributed. Hence, it generates an explanation for $\hat{y}$ as $(G_S, X_S^F)$, where $G_S$ is a sub-graph of the computation graph and $X_S^F$ are the most important features for $\hat{y}$. Thereby, a vital component of \emph{GNNExplainer} is learning real-valued graph- and feature-masks $M$, which is universal for all graph-based ML tasks. 
\subsubsection{Surrogate methods}
Approximating the predictions of a black box model by deploying a more simple yet interpretable model is referred to as a surrogate method. The surrogate model is a simpler and interpretable model that is easier to understand than the original model.
\citet{Zhang2021} addresses the underlying assumption \emph{GNNExplainer} to have access to the gradients of a model to learn explanations. The authors introduce \emph{RelEx}, a method that works with relation extraction that can handle complex and cross-sentence relations. The authors lever both syntactic and semantic information to extract relations from text. Their method remains model-agnostic and does not rely on the iid assumption. Most importantly, \emph{RelEx} resolves the limitation to differentiable relational models and therefore enhances practicality, compared to \emph{GNNExplainer}.\\
Another representative from the field of model-agnostic GNN explanation methods is \citet{Vu2020}. Referred to as \emph{Probabilistic Graphical Model-explainer}, the authors' manage to identify pivotal graph components and to deduce explanations in the form of a graphical model and conditional probabilities. Thereby, \emph{Probabilistic Graphical Model-explainer} use graph-based representation to encode complex distributions over a multidimensional space. Compared to \emph{GNNExplainer}, \emph{Probabilistic Graphical Model-explainer} would provide an explanation in the form of a \emph{Bayesian} network that estimates the probability that a  vertex $E$ has the predicted role given a realization of other vertices. \\
\begin{figure*}[!ht]
  \centering
  \includegraphics[width=\textwidth]{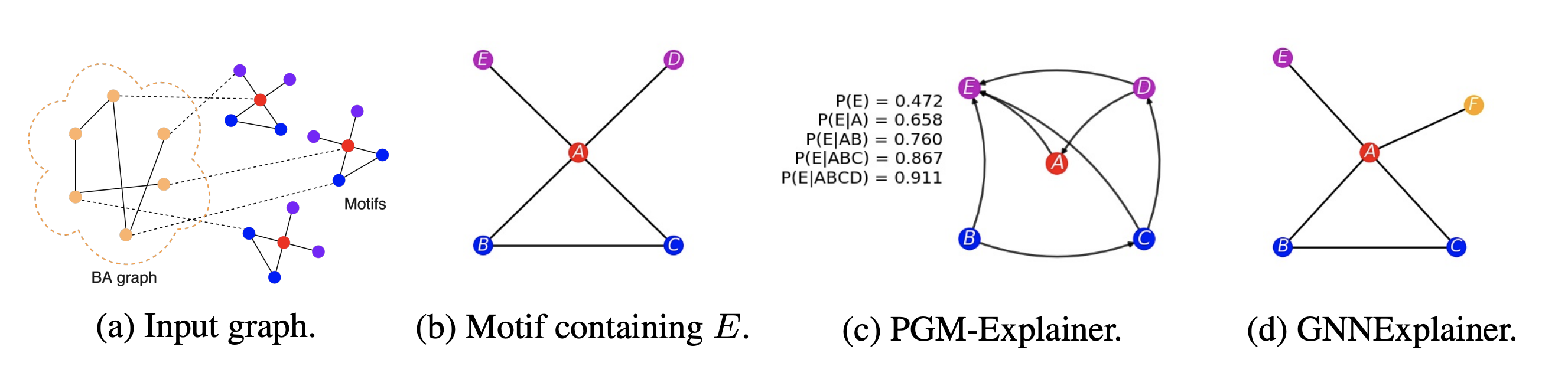}
  \caption{Exemplary \emph{Probabilistic Graphical Model-explainer} framework compared to \emph{GNNExplainer}, taken from \citep{Vu2020}.}\label{fig:pgm}
\end{figure*}
\emph{Probabilistic Graphical Model-explainer} is, as well, model-agnostic and thus, can be applied to any graph-based AI model, independent of its architecture. This epitomizes a crucial advantage, especially when it comes to practical application. Furthermore, the model is applicable to other graphical models such as dependency networks or \emph{Markov} networks.
One may argue that local explanations, such as provided by \citet{Vu2020}, lack generalizability for the whole graph. \emph{PGExplainer}, by \citet{Luo2020}, addresses this issue by parameterizing the process of explanation-generation, driven by the idea that neural network parameters are the same throughout the population. The intuition of the idea stems from conventional forward propagation-based methods and tries to identify a subgraph that contains explanatory vertices. \\
Other methods stress the proximity of a vertex of a KG, whose prediction is to be explained. In 2020, \citet{Huang2020} proposed \emph{GraphLIME} to explain any vertex by a set of features. Again, the method perturbs the input graph data around a specific instance of interest to create a modified graph. Next, the authors define a set of neighbours of a vertex, by the vertices whose distance from the is within a finite number of steps, frequently referred to as hops. Eventually, the nonlinear, feature-wise kernelized \emph{Hilbert-Schmidt} Independence Criterion (HSIC) Lasso  method is used to determine the least redundant feature vectors. This method is a supervised feature selection method and can be revised in \citet{Yamada2016}. Essentially, the framework proposed by the authors learns a non-linear interpretable model in the local proximity of a predicted vertex. As opposed to \emph{LIME}, a linear explanation model by \citet{Ribeiro2016}, \emph{GraphLIME} takes the graph topology into account. Experimenting with the real-world datasets \emph{CORA} and \emph{Pubmed}, the explanations of \emph{GraphLIME} are depicted as more descriptive, when compared to \emph{GNNExplainer}. \\
\begin{figure*}[!ht]
      \centering
      \includegraphics[width=.6\textwidth]{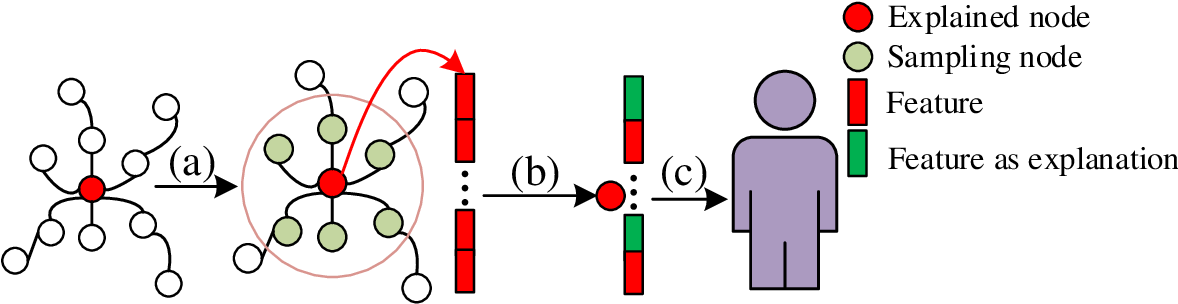}
      \caption{Generic components of \emph{GraphLIME}, taken from \citet{Huang2020}. The system samples neighbors of the node that is to be explained (in red) and vertices in green, which are in the proximity of the vertex that is to be explained (a). From the feature vectors in red, the explanatory parts are determined (b), and presented to the user (c).}\label{fig:glm}
\end{figure*}
Figure~\ref{fig:glm} displays the generic components of \emph{GraphLIME}, with a vertex in red, whose prediction is to be explained, vertices in green, which are in the proximity of the vertex that is to be explained (a). From the feature vectors in red, the explanatory parts are determined (b), and presented to the user (c).
On the vertex-level, \emph{GraphSVX} by \citet{Duval2021} constructs a surrogate model on a perturbed dataset. Then, \emph{GraphSVX} uses the surrogate model to compute the marginal contribution of vertices and features using the surrogate model binary masks $(M_N, M_F)$. Eventually, \emph{GraphSVX} draws on the so-called \emph{Shapely} value, a concept from game theory, which represents the fair distribution of gains in a game. Put differently, \emph{Shapely} values are a metric of the marginal contribution of each feature of a model's prediction. The authors approximate this metric in order to create a signification of an explanation obtained by \emph{GraphSVX}. Additionally, the authors describe a unified framework of GNN explanations, consisting of a mask generator, a graph generator and an explanation generator, in which they locate competing models, such as \emph{GNNExplainer}, \emph{PGExplainer}, \emph{PGM-Explainer} or \emph{GraphLIME}. Even though the authors use a super-ordinate graph for generating explanations, the scope of \emph{GraphSVX} is limited to vertex level predictions. 
\emph{GraphSHAP} however, introduced by \citet{Perotti2022} jointly makes use of the \emph{Shapely} value, but on a GC level. The system decomposes the graph into motifs, whereby motifs are recurring patterns in the KG. For instance, a motif could be a triangle, a four-clique, or a cycle. Thus, \emph{GraphSHAP} uses \emph{Shapely} values to calculate the contribution of each motif to the graph classifier's output.
Similarly, \citet{schlichtkrull2021interpreting} learn a straightforward classifier that predicts if an edge can be neglected or not, given its contribution to the prediction. For every node in every layer of a given, trained GNN, the system learns a classifier that predicts if that edge can be dropped. The authors claim to train the classifier in a fully differentiable fashion, employing stochastic gates and encouraging sparsity through the expected $L_0$ norm, which makes the classifier also interpretable attribution method to analyze Graph Neural Networks.
\subsubsection{Graph generation methods}
Some methods focus on training a graph generator instead of optimizing the input KG directly.
For GC tasks, \citet{Yuan2020XGNNTM} create a \emph{XGNN}, a method to explain GNN predictions on the model-level. The authors train a separate graph generator in a RL setting that predicts a suitable way to add a new edge to the input graph. The generator produces a perturbation of the input graph, whereby changes are made in a way that affects the GNN's prediction. The RL policy works with manually generate graph rules that eventually explain the contributions of both nodes and edges. \emph{XGNN} is appropriate for both graphs and KGs. 
Yet another model-agnostic but symbolic post-hoc explanation method for graph classifiers was proposed by \citet{Abrate2021} in 2021. The authors construct a counterfactual graph, which can, compared to \emph{Shapely} values, provide a deterministic explanation of why a classifier made a prediction. The method proposed by the authors is a heuristic search scheme that can be conducted in two variants: (1) Oblivious, the algorithm searches for the optimal counterfactual graph by querying the black-box graph classifier and abstracting the original graph. (2) Data-driven, the algorithm gains additional access to a subset, superset or the original training dataset of the black box classifier. Similar to \emph{SHAP} \citep{Lundberg2017} and \emph{LIME} \citep{Ribeiro2016}, the authors rank statistics about the input data in order to achieve explainability. However, they do so by creating numerous counterfactual graphs and eventually stressing the most frequent edges. Thus, the method of \citet{Abrate2021} creates counterfactual graphs for different input data as an explanation, resulting in a global and model-agnostic XAI method. The authors forgo on integrating higher-order structures or graph properties into their method.
A major issue for XAI in KG-based models is the occurrence of cyclic relations in the KG. Cycles occur when edges relate vertices in contradictory hierarchies
.
The main issue with a KG cycle is that reasoning over the graph in the form of choosing parent concepts within the KG hierarchy for generating explanations is rendered nondeterministic \citep{Sarker2020}. 
\citet{Sarker2020} address the issue of cycles within a KG by providing a circle-free version of the existing \emph{Wikipedia} KG (\emph{WKG}). In the \emph{WKG}, entities are articles in the English \emph{Wikipedia} and relations are hyperlinks between them. The authors also introduce a type system, relying on the \emph{Wikipedia} categories. Using breadth-first search, the authors break cyclic relations within the KG and thus create the so-called \emph{Wikipedia} KG. Interestingly, the authors introduce a measure for the XAI-related quality of their KG, using an ILP-based metric, focusing on string similarity, which shall not be detailed here but can be revised in the original paper (\citep{Sarker2020}) in conjunction with \citet{Levenshtein75}. 
Thus, explanations are tried to be extracted from existing KGs with cyclic relations, such as \emph{DBpedia}. Eventually, the \emph{WKG} can be used to identify the entities and types that are most relevant to a particular prediction and also to generate counterfactual explanations. Finally, the authors stipulate higher efficiency whilst minimal information loss by breaking cyclic relations.
Other methods based on the exploitation of semantic information exist. \citet{Dragoni2022} for instance, constructs a logical reasoning flow, referred to as explanation graph. The graph fulfills the formal criteria of a KG, as defined in Section~\ref{sub:meth}, but differs from previously described methods since it maps concepts used by users to numeric features of a black box model. In the end, the graph is a representation of the data that was used by an AI model for its prediction, being matched to given KG concepts. In fact, the authors aim at rendering an explanation in logical language format. For that, they create a framework being able to process semantic information from a knowledge base, referred to as semantic features $A_i$, based on the work of \citep{Doran2018}. In brief, semantic features represent the common attributes of an object, such as $3-Series \hookleftarrow (Car, BMW)$ and essentially are one kind of predicates of a FOL expression. The other kind of predicate are the output symbols of a black box model $O_i$ that are to be rendered comprehensible. Hence, being graph-based, the authors define their system as consisting of unary predicates $O_i, A_i$ as vertices and the logic relations between those predicates, coming from an AI system, as edges. The explanation graph hence combines singular knowledge fragments into a semantic reasoning system and thus creates explainability.
\citet{betz2022} propose a method based on adversarial attacks on KG embedding models. Adversarial attacks involve the perturbation of data during training, which results in model malfunction during testing. The authors leverage an efficient rule-learning technique, employing abductive reasoning to uncover triples that serve as logical justifications for specific predictions. Subsequently, the proposed attack centers around a straightforward concept: modifying or suppressing a triple within the most confident explanation. Remarkably, despite being independent of the model itself and requiring solely access to training data, the authors report outcomes on par with cutting-edge white-box attack techniques, which necessitate complete access to model architecture, learned embeddings, and loss functions. This surprising finding suggests that symbolic methods can partially retroactively elucidate KG embedding models. \\
Lastly, \emph{approxSemanticCrossE}, presented by \citet{dAmato2022} uses semantic similarity to identify the most important entities and relations that contributed to the prediction. The technique employs semantic similarity to discern the pivotal entities and relationships influencing predictions. \emph{approxSemanticCrossE} entails a three-step procedure: (1) Recognition of similar entities (i.e. the system identifies entities bearing the greatest resemblance to those involved in the link prediction, utilizing a semantic similarity metric). (2) Detection of similar relationships (i.e. \emph{approxSemanticCrossE} pinpoints relationships that closely resemble the predicted link, employing a semantic similarity measure). (3) Exposition generation: The final step encompasses generating an exposition, which manifests as a compilation of the most similar entities and relations, accompanied by their respective semantic similarity scores.
After having outlined the state of the art of XAI on KGs, Table~\ref{tab:literature_table_xai} displays the results of the literature review for this section.

\begin{table*}[!ht]
\centering
    \begin{tabular}{ | p{5.25cm} | p{4.4cm} | p{2cm}  p{1cm}  p{1.6cm} |}
    \toprule
    {\textbf{Line of research}}                                                 & {\textbf{References}}                                 & {\textbf{Representation}}       &  {\textbf{Task}}   & {\textbf{Foundation}}   \\
    \toprule

    \textbf{Rule-based}                                                         &                                                       &                                 &                    &  \\
    
    \small{\makecell[l]{Using symbolic representation to \\
    extract First Order Logic rules from KGs.}}                            & \emph{SeXAI} \citep{Donadello2021}                    & SB                              & Other               & RBL \\    
    
    \hline
    
    \small {Extracting FOL rules using the terminology of KGs.}            & Rule-based explanations \citep{Dervakos2022}          & SB                              & C                   & RBL \\
    
    \hline
    
    \small{Applying rule mining methods on embedding-based KG clustering.} & \emph{ExCut} \citep{Gad-Elrad2020}                    & SSB                             & C                   & RBL \\
    
    \toprule
    
    \textbf{Decomposition}                                                      &                                                       &                               &                     & \\
    
    \small{Identifying groups of relevant edges to construct relevant walks.}   & GNN LRP \citep{Bach2015, Montavon2019}  & SSB                    & Other        & ANN \\
    
    \toprule
    
    \textbf{Surrogate}                                                          &                                                       &                                &                    & \\
    
    \multirow{2}{*}{\small{\makecell[l]{
    Perturbing the input graph \\ 
    under IID assumptions.}}}                                              & \emph{GNNExplainer} \citep{NEURIPS2019_d80b7040}     & \multirow{2}{*}{SSB}   & \multirow{2}{*}{NC}    & \multirow{2}{*}{ANN} \\
                                                                                & \emph{RelEx} \citep{Zhang2021}                        &                               &                               &     \\
    
    \hline
    
    \multirow{2}{*}{\small{\makecell[l]{
    Using rules to create (graphical) explana- \\ 
    tions from relevant KG components.}}}                                  & \emph{Probabilistic Graphical Model-explainer} \citep{Vu2020}  & \multirow{2}{*}{SSB}   & \multirow{2}{*}{NC,GC} & \multirow{2}{*}{ANN} \\
                                                                                & \emph{PGExplainer} \citep{Luo2020}        &                               &                                           &                              \\
    \hline
    \small{Explaining KG vertices by a set of explanatory features.}       & \emph{GraphLIME} \citep{Huang2020}        & SSB                    & NC                                 & ANN                      \\ 
    \hline
    \multirow{3}{*}{\small{\makecell[l]{Using Shapely values or attribution.}}} & \emph{GraphSHAP} \citep{Perotti2022}     & \multirow{3}{*}{SSB}   & \multirow{3}{*}{NC, GC}     & \multirow{3}{*}{TL}      \\
                                                                                & \emph{GraphSVX} \citep{Duval2021}        &                               &                                           &                               \\ 
                                                                                & Diff. edge masking \citep{schlichtkrull2021interpreting}     &                               &                                           &                               \\
    \toprule
    \textbf{Graph generation}                                                   &                                                       &                               &                                           &                               \\                                                
    \small{Generating graphs by reinforced input graph perturbation.}                  & \emph{XGNN} \citep{Yuan2020XGNNTM}                    & SB                     & R                                  & RL                       \\
    \hline
    \multirow{2}{*}{\small{\makecell[l]{
    Generating graphs with \\ 
    counterfactuals and inductive logic.}}}                                     & Counterfactual graph \citep{Abrate2021}  & \multirow{2}{*}{SB}    & \multirow{2}{*}{GC}                & \multirow{2}{*}{Other}        \\
                                                                                & \emph{WKG} \citep{Sarker2020}                       &                               &                                           &                               \\
    \hline
    \multirow{2}{*}{\small{\makecell[l]{
    Generating graphs by matching \\ 
    concepts of a KG.}}}                                                   & Explanation graph \citep{Dragoni2022} & \multirow{2}{*}{SB}    & \multirow{2}{*}{Other}                    & \multirow{2}{*}{Other}        \\
                                                                                &                                      &                               &                                           &                               \\
    \hline
    \multirow{2}{*}{\small{\makecell[l]{
    Generating graphs / embeddings using \\ 
    GANs and semantic similarity.}}}                                            & Adversarial Explanations  \citep{betz2022} & \multirow{2}{*}{SB}    & \multirow{2}{*}{Other}                & \multirow{2}{*}{TL}      \\ 
                                                                                & \emph{approxSemanticCrossE} \citep{dAmato2022}      &                               &                                           &                               \\                         
    \bottomrule
    \end{tabular}
    \normalsize
    \caption{Literature review table of XAI on KGs. Each item is classified by representation, task, foundational method, and line of research. Sometimes, approaches are constructed by multiple publications building upon each other. In this case, all related publications are cited.}\label{tab:literature_table_xai}
\end{table*}

\begin{figure*}
    \centering
    \includegraphics[width=\textwidth]{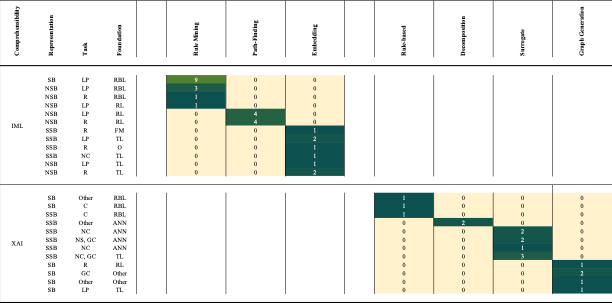}
    \caption{A heatmap of literature review results by representation, task and foundation. IML, XAI, and the determined lines of research are displayed separately.}
    \label{fig:heatmpa}
\end{figure*}
Additionally, \ref{fig:heatmpa} combines the taxonomy provided by this survey, summarized in the table above, with the discrimination of IML and XAI on KGs. The figure shows a heatmap of literature reviewed in this survey, displaying blank spots in current literature.
The following section shall conclude the literature review and provide an outlook for future research in the field of KGs for CAI.
\section{Conclusion}\label{sec:conc}
So far, we contributed to the research fields of XAI and IML by introducing the parent concept of CAI as well as a taxonomy for existing research in the conjunction of CAI with KGs. Now, the comprehensive overview of the preceding section shall be concluded before future research directions are described.

This survey provided a brief history of CAI on KGs before describing the proceeding of the survey. 
Next, we taxonimized CAI on KGs by the representation (cf. Section \ref{sub:represenation}), the task (cf. Section \ref{sub:tasks}), the foundational method (cf. Section \ref{sub:optimization}), and the kind of CAI (cf. Section \ref{sub:comp}), which we refer to as either IML or XAI. 
We provided a detailed notion of our scope in the methodology and definitions section. Additionally, we made a case for the necessity of CAI on KGs, defending the concept CAI, XAI and IML on KGs. We divided the main body of this survey into an IML and a XAI part, where we described 55 selected titles from our survey. We can conclude that a significant amount of research works with a sub-symbolic representation of KGs. 
Neural network-based methods are strongly represented in both XAI and IML on KGs. Meanwhile, advances using RL are made. Figure~\ref{fig:heatmpa} summarizes our results.
The following section shall provide the reader with future research directions of CAI on KGs and concludes this survey.

\paragraph{Outlook and future research directions}
    
Our survey unveiled that few XAI methods tackle the link prediction tasks and, at the same time, IML methods occur less frequently for vertex and graph classification. Thus, exploring XAI methods for link prediction is a research frontier with significant opportunities. For example, how can the decision process of a black box embedding model be reconstructed?  \\
In addition, most XAI methods discussed in this paper are not KG specific. They do not leverage the semantic information of the node labels, edge labels, edge directions, or underlying ontology. Future XAI on KGs should leverage this information to enrich the explanations, leading to significant insides in the black box models' behavior. \\
This survey also recognizes that the comparative evaluation of the XAI methods on KGs needs to be clarified. There is yet to be an established ground for comparing the performance of different XAI methods on KG. Questions like which XAI method reflects the models' behavior more accurately are regularly unanswered. Future research has to establish common evaluation grounds in the form of standardized evaluation metrics.  \\
At the same time, IML methods occur less frequently in vertex and graph clustering. Thus, IML methods for graph clustering hold significant potential for future research, as they promise to make graph clustering more performance and trustworthy.
For example, an interpretable rule-mining method could learn rules from the KG structure that cluster similar KGs in the same group. Similarly, vertex clustering algorithms could be assessed and enriched with embedding-based IML methods, such as \emph{CrossE} \citep{dAmato2022}. \\
Furthermore, we see significant potential in improving the communication of the IML model's interpretation. 
While they are white-box models, the interpretation of their decision-making process is seldom user-friendly and requires expert ML knowledge. This survey encourages future research on IML on KGs to provide user-friendly communication strategies for the model's interpretation via stakeholder-specific interfaces. For example, an interface could visualize the model's decision-making process by embedding it in a context provided by the knowledge graph for non-expert users. \\
AI applications are moving outside the safe walls of labs and are invading our daily lives. Let us use the semantics provided by KGs to make AI applications safer. 

\section*{Acknowledgments}
We received numerous e-mails with feedback on our survey and would like to thank all researchers that contributed with their comments and constructive criticism to this paper. We especially would like to thank Judith Wewerka for her in-depth feedback on our work.
\newpage
%



\bibliographystyle{elsarticle-num-names} 
\bibliography{cite}






-------------------------------------------
\newpage

\appendix
\section{Selected questionnaire answers.}\label{sec:appendix_a}
\begin{table}[!ht]
\centering
\tiny
    \begin{tabular}{p{0.4\textwidth}}
    \toprule
    \textbf{Which papers do you think are crucial to cite in such a survey?}\\
    \emph{
    To the best of my knowledge, very few papers tackled the explainability
    of knowledge graphs. The explainability of graphs was tackled by several
    works. [...] Although these approaches
    generalise well to KG, I believe there is some space for more targeted
    approach, which leverage the different types of edges / atoms more 
    explicitly into their functioning, proposing more balanced or type-specific
    explanations using these additional information.} \\
    \toprule
    \textbf{Which major outlets (conferences/journals) have you seen published
    literature on Comprehensible Artificial Intelligence on Knowledge Graphs?}\\
    \emph{
    [Multiple answers, all of which were integrated into this survey.]} \\
    \toprule
    \textbf{Are there topics you would like to see discussed in such a survey?}\\
    \emph{
    I think real-world comprehensibility entails a mixture of / a continuum
    between explanation of the mechanics of the model and
    explanation of the world. In non-trivial knowledge domains,
    explanations will likely include information that is novel to the user
    and that not only describes how the model worked, they will also need to be shown
    and explained the world-knowledge that is relevant, and need to judge
    the relevance and validity of that knowledge.} \\
    \toprule
    \textbf{Where do you see future research directions for Comprehensible Artificial Intelligence on Knowledge Graphs?}\\
    \emph{
    I have the suspicion that approaches existing approaches [...]
    are close to optimal when it comes to purely graph-structured data.
    I think we will see increasing inclusion of additional, non-graph data
    (especially text data processed by language models) in order to go beyond the current state-of-the-art of predictive performance. [...]} \\
    \bottomrule
    \end{tabular}
    \normalsize
\end{table}

\end{document}